\documentclass[10pt,twocolumn,letterpaper]{article}

\usepackage{cvpr}              

\usepackage{caption}

\definecolor{cvprblue}{rgb}{0.21,0.49,0.74}
\usepackage[pagebackref,breaklinks,colorlinks,allcolors=cvprblue]{hyperref}

\def\paperID{} 
\def\confName{CVPR}
\def\confYear{2026}

\title{MoRe: Motion-aware Feed-forward 4D Reconstruction Transformer}

\author{
    Junton Fang$^{1*}$, Zequn Chen$^{2*}$, Weiqi Zhang$^{1*}$, \\
    Donglin Di$^2$, Xuancheng Zhang$^{1,2}$, Chengmin Yang$^2$, Yu-Shen Liu$^{1\dagger}$ \\
    $^1$School of Software, Tsinghua University, Beijing, China \qquad $^2$Li Auto \\
    {\tt\small fangjt21@mails.tsinghua.edu.cn, chenzequn@lixiang.com, zwq23@mails.tsinghua.edu.cn} \\
    {\tt\small \{didonglin, zhangxuancheng, yangchengmin\}@lixiang.com, liuyushen@tsinghua.edu.cn}
}

\begin{document}
\twocolumn[{
    \maketitle
    \vspace{-1em}
    \begin{center}
        \includegraphics[width=\linewidth]{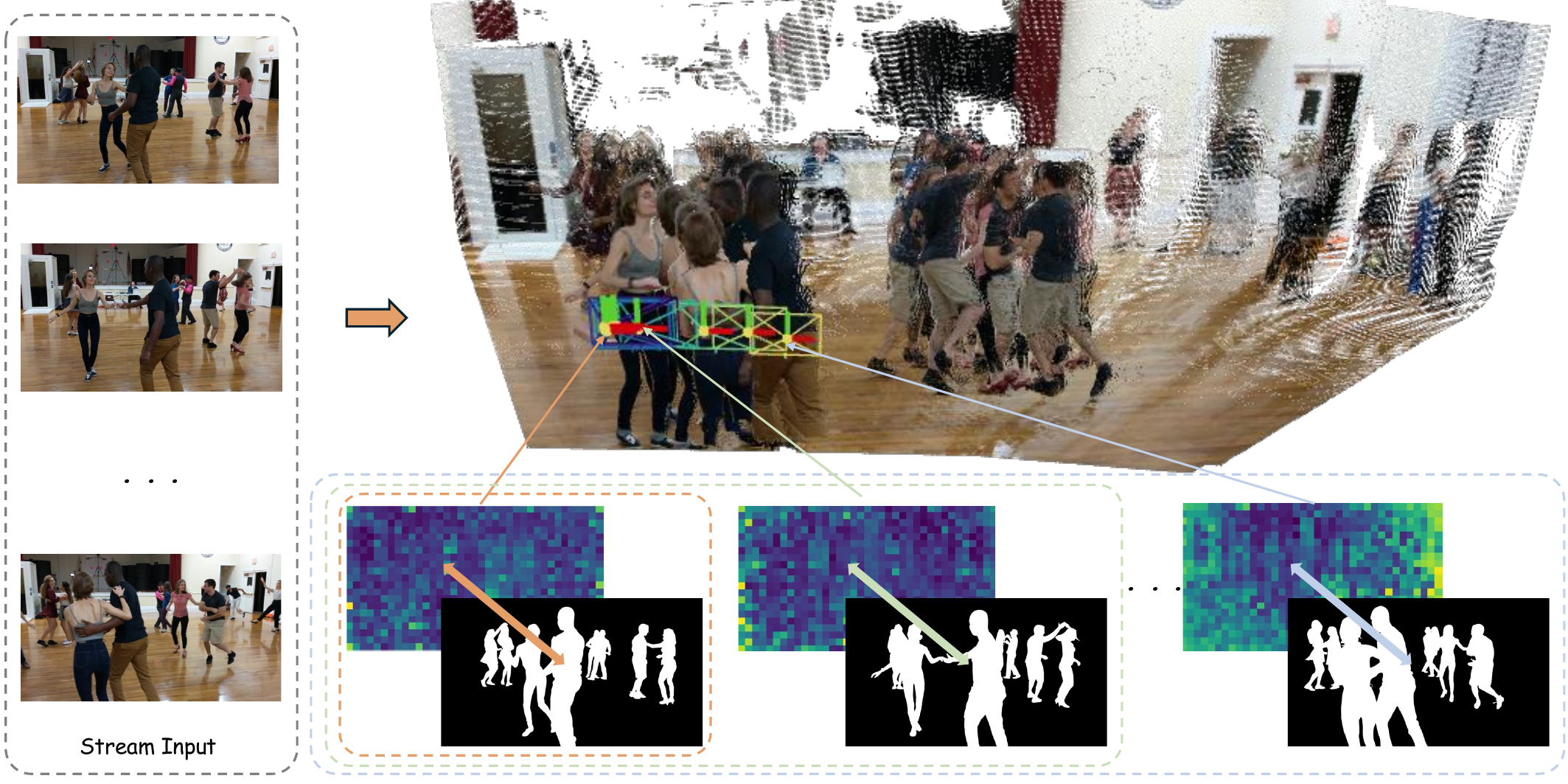}
        \vspace{-0.5em}
        \captionof{figure}{
        We propose MoRe, a motion-aware 4D reconstruction transformer that explicitly disentangles dynamic motion from static scene structure. This capability is enabled by our attention-forcing training strategy, which guides the model to separate motion cues from background geometry. At inference time, More further supports streaming inputs through its grouped causal attention design.}
        \label{fig:cover}
        \vspace{1em}
    \end{center}
}]
{
  \renewcommand{\thefootnote}{\fnsymbol{footnote}}
  \footnotetext[1]{Equal contribution, where Zequn leads this project.}
  \footnotetext[2]{The corresponding author is Yu-Shen Liu. This work was partially supported by Deep Earth Probe and Mineral Resources Exploration—National Science and Technology Major Project (2024ZD1003405), and the National Natural Science Foundation of China (62272263).}
}
\begin{abstract}
Reconstructing dynamic 4D scenes remains challenging due to the presence of moving objects that corrupt camera pose estimation. Existing optimization methods alleviate this issue with additional supervision, but they are mostly computationally expensive and impractical in real-time applications. To address these limitations, we propose MoRe, a feedforward 4D reconstruction network that efficiently recovers dynamic 3D scenes from monocular videos. Built upon a strong static reconstruction backbone, MoRe employs an attention-forcing strategy to disentangle dynamic motion from static structure. To further enhance robustness, we fine-tune the model on large-scale, diverse datasets encompassing both dynamic and static scenes. Moreover, our grouped causal attention captures temporal dependencies and adapts to varying token lengths across frames, ensuring temporally coherent geometry reconstruction. Extensive experiments on multiple benchmarks demonstrate that MoRe achieves high-quality dynamic reconstructions with exceptional efficiency. Project page: \url{https://hellexf.github.io/MoRe/}.
\end{abstract}    
\section{Introduction}
\label{sec:intro}
Reconstructing the evolving three-dimensional structure of a scene (i.e., 4D reconstruction) is increasingly central to applications in augmented reality, robotics, digital twins, and immersive content creation\cite{gaussiangrow, NEURIPS2024_41fb2ecb, Zhou_2024_CVPR, Zhang_2025_ICCV}. Classical geometry based techniques such as SfM/MVS~\cite{schonberger2016structure,furukawa2015multi} and SLAM~\cite{taketomi2017visual,davison2007monoslam,newcombe2015dynamicfusion} have laid the foundation by estimating camera poses and scene structure under the assumption of a mostly static environment. They achieve high accuracy in controlled settings but struggle when objects deform or move, or when the camera undergoes complex motion.

Recent deep learning methods promise faster and more generalizable reconstruction~\cite{wang2025vggt, keetha2025mapanything, wang2024dust3r, yang2025fast3r, dens3r, du2025superpc, yao2026anchoreddreamzeroshot360degindoor, MoRE2025, noda2024multipull, zhou2026udfstudio, Zhou_2024_CVPR, zhang2025materialrefgs}. They split broadly into real time inference models and hybrid optimization pipelines. Real time reconstruction models map image sequences directly to camera poses and depths (or point clouds) in one feedforward pass, enabling fast processing of video input. However, these approaches are trained predominantly on static scenes. The presence of moving objects or large camera motion can significantly degrade the accuracy of estimated 3D structure. Hybrid optimization pipelines~\cite{li2025megasam,yao2025uni4d,wang20243d,zhang2024monst3r} integrate learned modules such as depth estimation, optical flow estimation, or motion segmentation. However, they retain a multistage structure or rely on iterative refinement. These methods handle dynamic scenes more robustly but incur high computational cost and struggle when processing long sequences or streaming video data in real time.

A clear gap remains: how to design a fast, generalizable framework that handles camera and object motion in dynamic scenes under streaming or long sequence input while producing accurate camera poses and depths for point cloud reconstruction. We propose MoRe, a motion aware 4D streaming reconstruction system for monocular video. Our core innovation lies in teaching the reconstruction model to distinguish dynamic objects from the static background purely through training, without introducing explicit motion or segmentation priors during inference. MoRe builds upon a strong reconstruction backbone and introduces an attention forcing strategy that explicitly supervises motion disentanglement while implicitly preserving geometric consistency. This integration allows the network to learn how motion influences both scene structure and camera trajectory, enabling robust depth and pose estimation even under significant dynamic movement.

To enhance temporal coherence and scalability, we propose a temporally aware streaming inference strategy combining grouped causal attention with a bundle adjustment~\cite{hartley2003multiple} like incremental refinement process. The attention mechanism captures long range temporal dependencies and adapts to varying token lengths across frames. The refinement process incrementally updates camera poses and scene geometry to maintain temporal consistency. Combined with large scale finetuning on diverse static and dynamic datasets, MoRe achieves fast, motion aware, and generalizable 4D reconstruction suitable for real world dynamic environments.

We train our model end to end on large scale 3D datasets covering diverse static and dynamic scenarios and comprehensively evaluate its performance across multiple downstream applications. Our main contributions are as follows:
\begin{itemize}
    \item We present MoRe, a unified motion aware 4D reconstruction framework capable of jointly estimating camera poses, depths, and motion masks in dynamic scenes.
    \item We introduce an attention forcing strategy that effectively teaches the network to disentangle dynamic motion from static structure during training through explicit supervision and implicit geometric consistency.
    \item We design a temporally aware inference mechanism combining grouped causal attention and bundle adjustment like streaming refinement, which captures long range dependencies while performing lightweight global refinement.
    \item Extensive experiments on diverse benchmarks demonstrate that MoRe achieves state of the art accuracy and strong generalization in dynamic 4D reconstruction.
\end{itemize}
\section{Related Work}
    \subsection{4D Reconstruction}
    4D reconstruction recovers time-evolving 3D structures by jointly predicting camera poses and depth. Optimization-based systems~\cite{li2025megasam,yao2025uni4d} refine geometry and motion using auxiliary cues (e.g., optical flow, masks), but are computationally heavy for slong sequences or streaming input. Recent modular pipelines~\cite{lu2025align3r,zhang2024monst3r} leverage foundation models yet remain complex for real-time use. Another line of work~\cite{sucar2025dynamic,jin2025stereo4d,zhang2025pomato,feng2025st4rtrack} regresses temporally aligned point-maps; however, they often lack explicit motion-static disentanglement and require dense 4D supervision. Unlike these, MoRe preserves an elegant, easy-to-use feed-forward architecture while removing the need for auxiliary motion priors or extra annotations: we teach the model to separate dynamic and static regions during training so that inference remains lightweight, end-to-end and suitable for streaming video.
    \subsection{Learning-based Reconstrucion}
    \begin{figure*}[!h]
      \centering
      \vspace{-1em}
      \includegraphics[width=\linewidth]{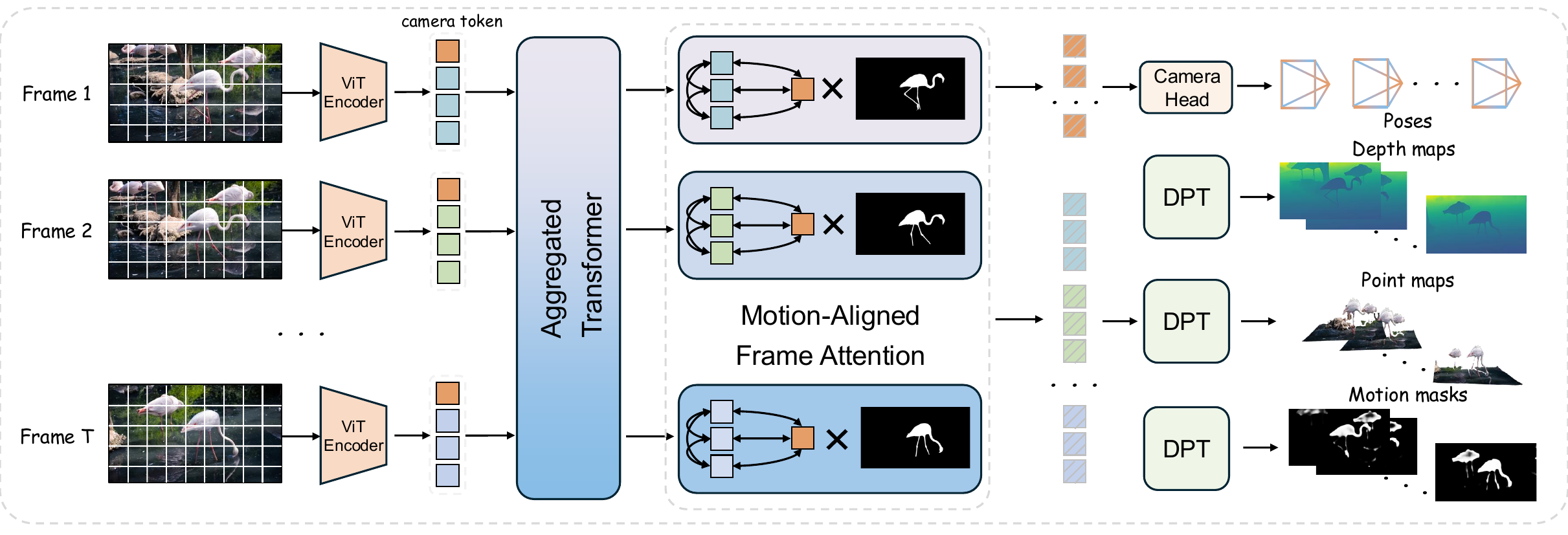} 
      \caption{\textbf{Method Overview.} During training, an attention-forcing mechanism aligns the attention weights with ground-truth motion masks, enabling the model to effectively disentangle dynamic motion from static scene structure. For streaming reconstruction task, MoRe is based on a causal transformer where global attention is replaced by aggregated causal attention. }
      \label{fig:pipeline}
      \vspace{-1em}
    \end{figure*}
    Recent learning-based methods directly regress geometry from images. Dust3R~\cite{wang2024dust3r} formulates reconstruction as point-map regression, eliminating explicit calibration. Subsequent works like MASt3R~\cite{leroy2024grounding} and Fast3R~\cite{yang2025fast3r} improve correspondences and scalability, while VGGT~\cite{wang2025vggt} uses a large Transformer to infer poses and depth maps across many views. Despite these advances, Transformer-based solutions face two limitations: quadratic inference costs with respect to sequence length, making streaming impractical, and a lack of explicit motion modeling, which reduces robustness in dynamic scenes.Despite their advances, Transformer-based solutions still face two key limitations: inference cost often scales quadratically with input length (making long or streaming video sequences impractical), and they typically neglect explicit modelling of camera or scene motion—which reduces robustness when objects or the camera itself move significantly. Our method addresses these shortcomings by introducing a causal attention mechanism and a streaming inference strategy, enabling efficient long-sequence processing and motion-aware geometry estimation.
    \subsection{Streaming Reconstruction}
    Sreaming reconstruction in dynamic scenes builds on visual SLAM, which estimates trajectories and maps incrementally. However, SLAM typically assumes static environments, limiting its 4D applicability. Recent frameworks like CUT3R~\cite{wang2025continuous} introduce transformer-based persistent latent states for online dense reconstruction. Following this, several systems~\cite{zhuo2025streaming,lan2025stream3r,li2025wint3r} adopt LLM-style architectures~\cite{brown2020language,touvron2023llama} and unidirectional causal attention with KV-caching to handle long sequences. However, standard LLM causal attention in 3D reconstruction faces two issues: first, it breaks intra-frame token correspondences by treating tokens as a flat sequence; second, streaming input imposes hard constraints where errors accumulate and long-term context drift persists. In contrast, our method integrates a causal attention mechanism tailored for image-tokens, combined with a BA-like token aggregation mechanism for global refinement.
\section{Method}
To address disconnections in 4D reconstruction, we propose MoRe, a feed-forward transformer designed for streaming input with a BA-like global alignment module. As shown in \cref{fig:pipeline}, MoRe jointly predicts per-frame depths, camera poses, point maps, and motion masks from monocular video. Our approach is built upon a formal problem formulation (\cref{subsec:problem-formulation}), followed by an attention-forcing strategy (\cref{subsec:motion-aligned-attn}) that integrates implicit and explicit supervision for temporal consistency. For efficient inference, we introduce a frame-aware grouped causal attention mechanism (\cref{subsec:grouped-causal-attn}) tailored for streaming reconstruction, with the overall training objective detailed in \cref{subsec:training-objective}.
    \subsection{Problem Formulation}
    \label{subsec:problem-formulation}
    Given a monocular video consisting of a frame sequence $\{I_t \in \mathbb{R}^{3 \times H \times W}\}_{t=1}^T$, our objective is to jointly estimate the per-frame depths $\{D_t \in \mathbb{R}^{H \times W}\}_{t=1}^T$, camera parameters $\{g_t \in \mathbb{R}^9\}_{t=1}^T$, and dynamic point maps $\{P_t \in \mathbb{R}^{3 \times H \times W}\}_{t=1}^T$.
    
    4D reconstruction typically involves long temporal sequences that arrive continuously and must be processed in real time. To accommodate this streaming nature, we reformulate the task as an online process:
    \begin{equation}
    \{D_t, g_t, P_t\}_{t=1}^T\ = f_\theta(\{C_t\}_{t=1}^{T-1}, I_{T-1}),
    \end{equation}
    where $C_t$ represents the cached information for $I_t$.
    
    Yet, streaming reconstruction further amplifies the uncertainty introduced by dynamic regions, where object motion leads to inconsistent geometry and appearance over time. The priors used in existing approaches are mainly effective in static areas and tend to fail in dynamic ones. To address this issue, we introduce motion masks ${M_t}_{t=1}^T$ as auxiliary guidance for motion-aware reconstruction:
    \begin{equation}
    \{D_t, g_t, P_t, M_t\}_{t=1}^T\ = f_\theta(\{C_t\}_{t=1}^{T-1}, I_{T-1}).
    \end{equation}
    
    Notably, our method does not rely on motion masks as explicit inputs, since their ground truth is typically unavailable; instead, motion cues are implicitly inferred and integrated into the model’s representation.
    
    Building upon this formulation, we next describe how our attention-forcing strategy enables MoRe to effectively exploit temporal dependencies while remaining robust to motion-induced ambiguities.
    \subsection{Motion-aligned Attention}
    \label{subsec:motion-aligned-attn}
    \begin{figure}[hbtp]
      \centering
      \vspace{-1em}
      \includegraphics[width=\linewidth]{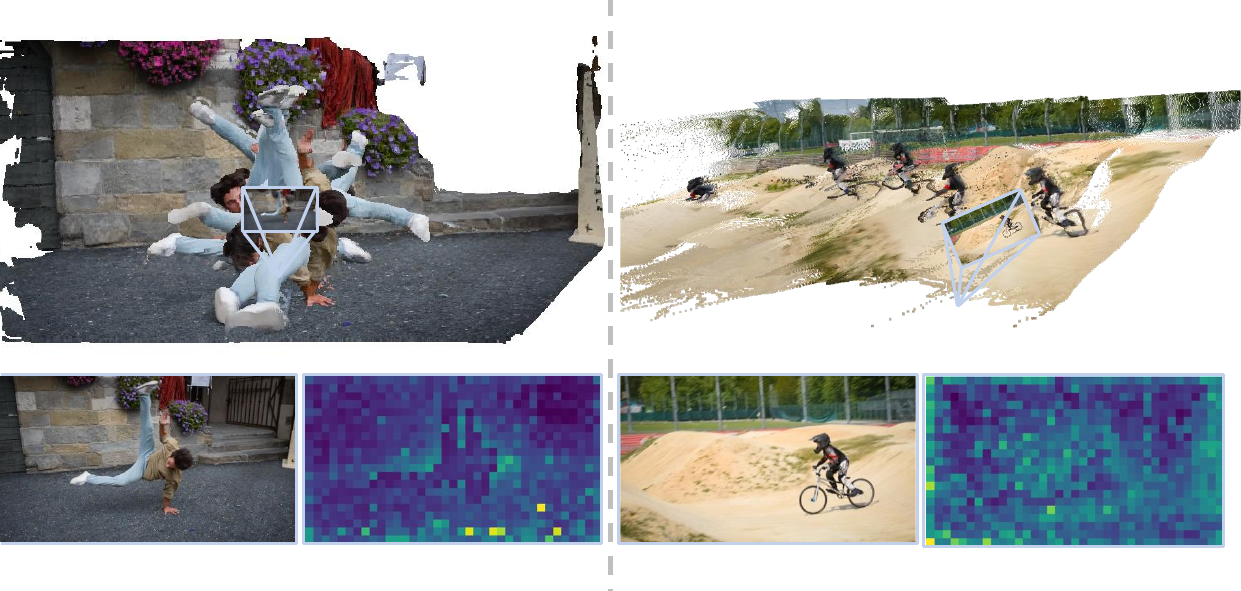} %
      \vspace{-2em}
      \caption{\textbf{Attention Map Visualization.} We visualize the attention map of the camera token within VGGT ~\cite{wang2025vggt} and observe that the model tends to confuse moving objects with static background regions, which accounts for the degradation in prediction accuracy. }
      \label{fig:attention map visualization}
      \vspace{-1em}
    \end{figure}
    In this section, we describe how our model disentangles dynamic motion from static structure. The key strategy relies on ground-truth motion masks during training and is completely test-time-free, avoiding any additional overhead during inference.

    Our motivation comes from directly transferring the foundation model to 4D reconstruction. As illustrated in \cref{fig:attention map visualization}, VGGT~\cite{wang2025vggt} performs well on the left example, where the camera token largely ignores moving regions. In contrast, in the right example, the camera token distributes nearly uniform attention across the image, indicating motion-induced confusion. When dynamic objects appear within an otherwise static scene, features used for camera estimation are severely corrupted, leading to degraded performance. This observation motivates the design of \textit{motion-aligned attention}, which explicitly guides the model to focus on static regions while partially ignoring moving objects.
    
    Motion-aligned attention is implemented by leveraging ground-truth motion masks during training. Given a motion mask $M_t$, we divide it into patches of size $s \times s$ consistent with image tokenization, producing mask tokens $\{m_i\}_{i=1}^{\frac{H}{s}\times\frac{W}{s}}$. The motion score for each image token is computed via average pooling over its corresponding mask token:
    \begin{equation}
    a_i = 1 - \frac{1}{s^2} \sum_{(u,v)\in m_i} m_i(u,v),
    \end{equation}
    where $a_i \in [0,1]$ represents the prior we have for image token $i$, with higher values corresponding to static regions. These motion scores provide a soft supervision signal, allowing the model to learn which regions should contribute more to camera estimation.
    
    Crucially, the camera token’s attention weights $\{\alpha_i\}$ over the image tokens can be interpreted as a probability distribution. Specifically, $a_i$ serves as a penalty prior to modulate the distribution of $\alpha_i$. By supervising $\alpha_i$ based on $a_i$, we provide \textit{explicit supervision} that guides the model to differentiate between static and dynamic regions, thereby improving robustness to motion. By relying solely on ground-truth masks during training, this method avoids introducing extra outputs or computations during inference, making it fully test-time-free and suitable for streaming or real-time 4D reconstruction scenarios.
    
    Overall, motion-aligned attention allows the model to selectively attend to informative static regions, mitigating the negative impact of dynamic objects and enabling more accurate and stable camera and scene reconstruction in challenging dynamic environments.

    \subsection{Grouped Causal Attention}
    \label{subsec:grouped-causal-attn}
    In this section, we mainly discuss the streaming inference mechanism that allows MoRe to incrementally reconstruct 4D scenes in real time, based on a specially designed \textit{grouped causal attention} scheme.
    \

    Causal attention has been widely adopted in large language models to enforce autoregressive dependency across tokens, ensuring that each token only attends to its past context. However, such a formulation is not directly suitable for our image-based reconstruction scenario. Image tokens within the same frame should maintain mutual visibility to preserve spatial coherence and consistent geometric reasoning. Therefore, we reformulate the conventional upper-triangular causal mask into a frame-wise causal mask, as illustrated in Figure, which enforces temporal causality across frames while allowing full bidirectional attention within each frame. This adaptation enables MoRe to simultaneously maintain causal temporal reasoning and spatial consistency.
    \begin{figure}
      \centering
      \includegraphics[width=\linewidth]{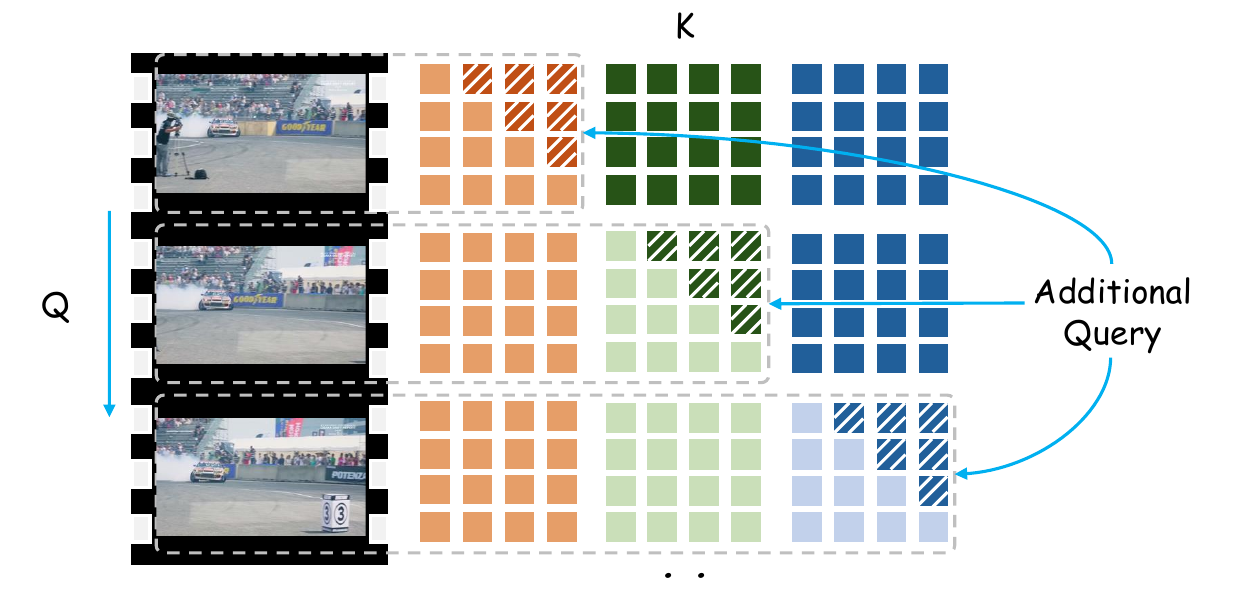} %
      \caption{\textbf{Grouped Causal Attention.} Unlike traditional causal attention, our design allows image tokens within the same frame to attend to each other regardless of their ordering. This formulation enables the model to preserve causal temporal reasoning while maintaining spatial consistency within each frame.}
      \vspace{-2em}
      \label{fig:causal attention}
    \end{figure}

    During streaming inference, the first image pair initializes the key--value (KV) cache. For each subsequent frame $I_t$, MoRe performs causal attention over the accumulated context from previous frames:
    \begin{equation}
    \begin{aligned}
        &F_t = \operatorname{Attn}\!\left(\mathbf{Q}_t, [\mathbf{K}_{1:t-1}, \mathbf{K}_t], [\mathbf{V}_{1:t-1}, \mathbf{V}_t]\right), \\
        &\mathrm{KV}_{1:t} \leftarrow [\mathrm{KV}_{1:t-1}; (\mathbf{K}_t, \mathbf{V}_t)],
    \end{aligned}
    \end{equation}
    Here, $F_t$ denotes the extracted feature representation for frame $I_t$, and $\mathrm{KV}_{1:t}$ stores all key–value pairs up to time $t$. This design enables MoRe to process frames sequentially while preserving temporal causality and avoiding redundant recomputation, leading to highly efficient streaming 4D reconstruction.

    However, such a strictly causal formulation also introduces a limitation: since each camera token only attends to its current and past contexts, long-term global information exchange becomes restricted. As a result, the accuracy of camera pose estimation may gradually degrade across extended sequences. To address this issue, we introduce a \textit{bundle-adjustment-like token aggregation} mechanism, which serves as a lightweight post-hoc refinement step after the streaming inference.
    
    Specifically, we cache the camera queries $\mathbf{Q}_t^{\text{cam}}$ during inference alongside the key--value features of all frames. Once the full sequence has been processed, each camera token performs an additional attention pass over all cached features to recover global geometric consistency:
    \begin{equation}
        \mathbf{C}_t^{\text{opt}} 
        = \operatorname{Attn}\!\left(\mathbf{Q}_t^{\text{cam}}, [\mathbf{K}_{1:T}], [\mathbf{V}_{1:T}]\right),
    \end{equation}
    This aggregation mechanism is analogous to the optimization step in bundle adjustment, effectively refining camera parameters in a globally consistent manner while maintaining real-time inference efficiency.
    
    Overall, the proposed streaming inference framework enables MoRe to achieve both efficiency and accuracy: causal attention ensures online processing capability, while the global token aggregation guarantees stable geometric reconstruction even in long sequences.
    \subsection{Training Objective}
    \label{subsec:training-objective}
    \begin{figure}
      \centering
      \includegraphics[width=\linewidth]{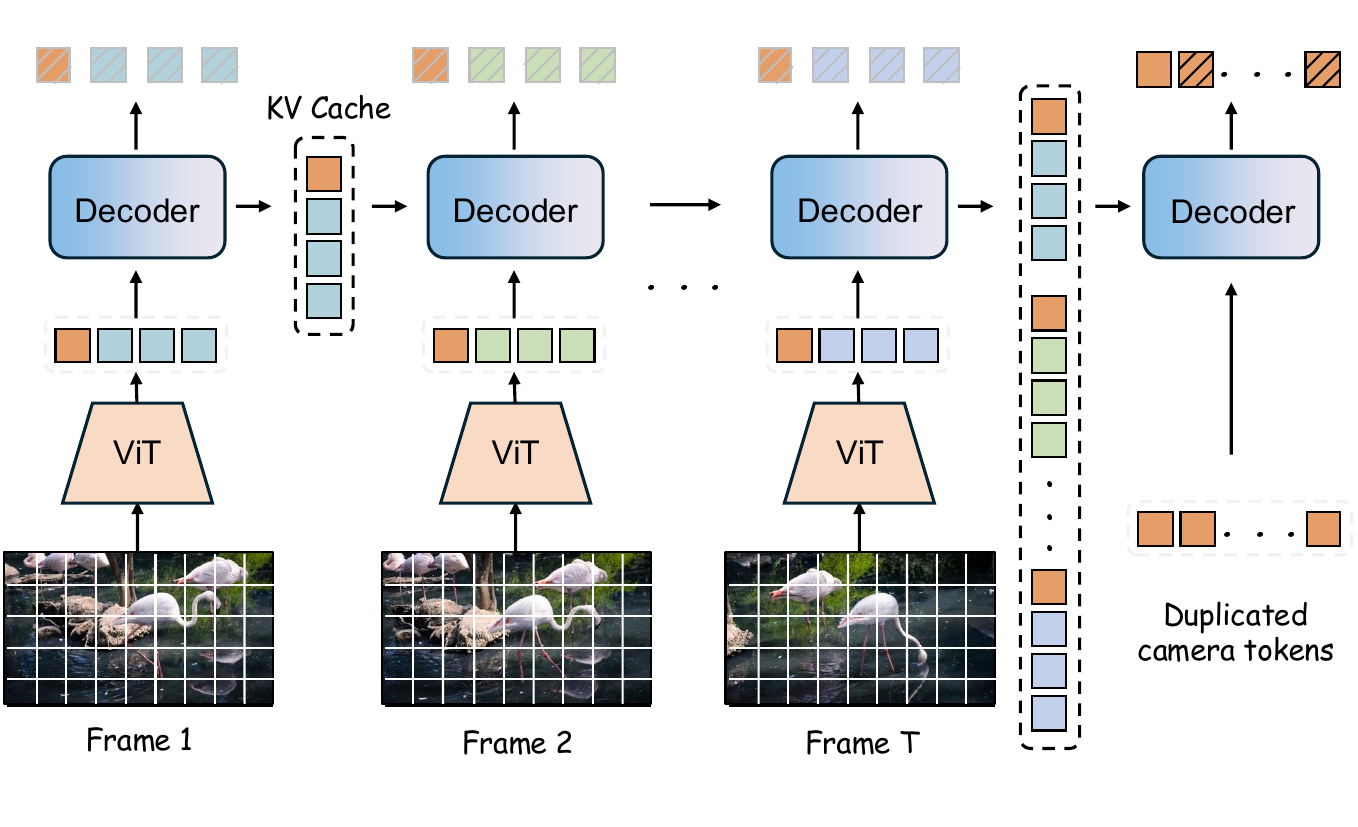} %
      \vspace{-2em}
      \caption{\textbf{Streaming Inference pipeline.} Leveraging causal attention, our model can efficiently process streaming input in an online manner. To enhance camera pose accuracy, we apply a bundle-adjustment-like post-processing step after the entire sequence has been processed. Specifically, for each frame, we duplicate the camera token and perform inference again using the previously cached key-value pairs.}
      \vspace{-2em}
      \label{fig:stream_mechanism}
    \end{figure}
    \begin{figure*}[!h]
      \centering
      \vspace{-1em}
      \includegraphics[width=\linewidth]{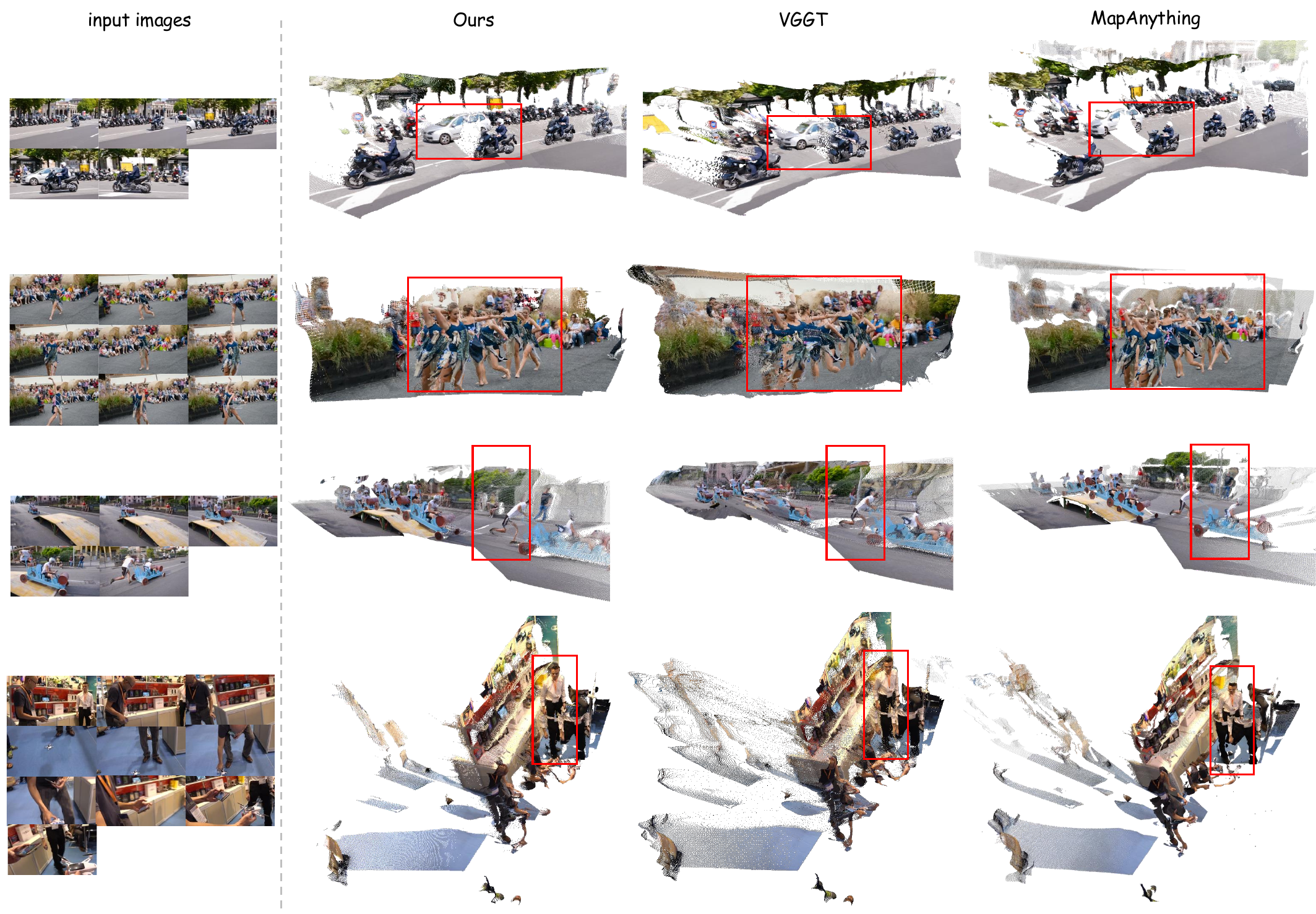}
      \caption{\textbf{Qualititive Comparison of Our Full Attention Model with Other Methods.} MoRe delivers outstanding performance in real-world scenes, outperforming other methods through its precise and robust geometry estimation.}
      \vspace{-1em}
      \label{fig:fa_visualization}
    \end{figure*}
    Following VGGT~\cite{wang2025vggt} and Dust3R~\cite{wang2024dust3r}, our training objective combines multi-task regression and classification, covering depth maps, point maps, camera parameters, and motion masks. For depth and point map regression, we adopt a confidence-weighted regression loss, which encourages both accurate predictions and reliable confidence estimates:
    \begin{equation}
    \mathcal{L}_{\text{conf}} = \sum_{i=1}^{N} 
    \Big(\hat{c}_i \, \| \hat{y}_i - y_i \|_2^2 - \lambda \log(\hat{c}_i)\Big),
    \end{equation}
    where $N$ is the number of pixels or points, $\hat{y}_i$ and $y_i$ denote the predicted and ground-truth values respectively, and $\hat{c}_i \in [1,\infty)$ represents the predicted confidence for each point.

    For motion mask prediction, we employ a standard binary cross-entropy (BCE) loss, supervising each pixel to indicate whether it belongs to a dynamic region:
    \begin{equation}
    \mathcal{L}_{\text{motion}} = - \frac{1}{N} \sum_{i=1}^{N} 
    \left[ M_i \log(\hat{M}_i) + (1 - M_i) \log(1 - \hat{M}_i) \right].
    \end{equation}

    As mentioned in Sec. \ref{subsec:motion-aligned-attn}, we encourage the camera token’s attention distribution over image tokens to align with the normalized motion scores. We use a simple confidence-like loss to enforce this alignment:
    \begin{equation}
    \mathcal{L}_{\text{attn}} = \frac{1}{M}\sum_{i=1}^M \max(0, a_i - C) \cdot \alpha_i.
    \end{equation}
    where $M$ is the number of image tokens and $C$ is a constant to control the penalty region. 

    To enhance both the accuracy of the streaming predictions and the final grouped camera estimates, we adopt a training strategy that explicitly supervises the camera token in two parallel paths. During training, for each frame, we duplicate the camera token and move them to the end of the sequence. Both the original and the duplicated tokens are decoded to predict camera parameters, which are then compared to the ground truth. This encourages the model to maintain consistent predictions across both the streaming path and the post-hoc aggregation path.

    \begin{table*}
    \centering
    \footnotesize
    \caption{Camera Pose Estimation on Sintel~\cite{butler2012naturalistic}, TUM-dynamics~\cite{sturm2012benchmark}, Bonn~\cite{palazzolo2019refusion}, and ScanNet~\cite{dai2017scannet} datasets. FA refers to full attention.}
    \setlength{\tabcolsep}{3pt}
    \label{tab:pose_estimation}
    \begin{tabular}{c c ccc|ccc|ccc|ccc}
    \toprule
     & &\multicolumn{3}{c}{\textbf{Sintel}} & \multicolumn{3}{c}{\textbf{Bonn}} & \multicolumn{3}{c}{\textbf{TUM-dynamics}} & \multicolumn{3}{c}{\textbf{ScanNet}(static)} \\
    \cmidrule(lr){3-5} \cmidrule(lr){6-8} \cmidrule(lr){9-11} \cmidrule(lr){12-14}
     \textbf{Method} & \textbf{type} & ATE$\downarrow$ & RPE$_\text{trans}$$\downarrow$ & RPE$_\text{rot}$$\downarrow$ & ATE$\downarrow$ & RPE$_\text{trans}$$\downarrow$ & RPE$_\text{rot}$$\downarrow$ & ATE$\downarrow$ & RPE$_\text{trans}$$\downarrow$ & RPE$_\text{rot}$$\downarrow$ & ATE$\downarrow$ & RPE$_\text{trans}$$\downarrow$ & RPE$_\text{rot}$$\downarrow$ \\
    \midrule
    MapAnything~\cite{keetha2025mapanything} & FA & 0.2104 & 0.0919 & 2.7396 & 0.0248 & 0.0132 & 0.6927 & 0.0244 & 0.0161 & 0.3871 & 0.0603 & 0.0280 & 0.8490\\
    VGGT~\cite{wang2025vggt} & FA & \underline{0.1715} & \underline{0.0617} & \underline{0.4695} & \underline{0.0141} & \underline{0.0100} & \underline{0.6323} & \textbf{0.0109} & \underline{0.0092} & \underline{0.2992} & \textbf{0.0347} & \underline{0.0151} & \textbf{0.3758}\\
    MoRe(FA) & FA & \textbf{0.0877} & \textbf{0.0580} & \textbf{0.3899} & \textbf{0.0138} & \textbf{0.0099} & \textbf{0.6267} & \underline{0.0115} & \textbf{0.0088} & \textbf{0.2980} & \underline{0.0375} & \textbf{0.0147} & \underline{0.3847}\\
    \midrule
    Spann3R~\cite{wang20243d} & Streaming & 0.3313 & 0.1111 & 4.4952 & 0.0344 & 0.0212 & 2.2539 & 0.0421 & 0.0333 & 0.8120 & 0.0651 & 0.0339 & 0.9348  \\
    CUT3R~\cite{wang2025continuous} & Streaming & 0.2163 & \textbf{0.0756} & \underline{0.6518} & 0.0420 & 0.0094 & 0.6825 & 0.0438 & 0.0134 & 0.4210 &	0.0929 & 0.0223 & \underline{0.5811}  \\
    StreamVGGT~\cite{zhuo2025streaming} & Streaming & 0.4159 & 0.1097 & 1.1056 & 0.0451 & 0.0148 & \textbf{0.5982} & 0.0760 & 0.0317 & 0.7012 & 0.1436 & 0.0437 & 1.6533    \\
    Wint3R~\cite{li2025wint3r} & Streaming & 0.2251 & 0.0972 & 1.0922 & 0.0366 & \textbf{0.0084} & 0.7170 & 0.0700 & 0.0221 & 0.7325 & 0.0618 & \textbf{0.0204} & 0.7028     \\
    Stream3R~\cite{lan2025stream3r} & Streaming & \underline{0.2144} & \underline{0.0764} & 0.8674 & \underline{0.0235} & \underline{0.0108} & 0.6664 & \textbf{0.0240} & \underline{0.0123} & \textbf{0.3180} & \textbf{0.0521} & 0.0215 & 0.9919   \\
    MoRe & Streaming &  \textbf{0.1474} & 0.0776 & \textbf{0.6157} & \textbf{0.0211} & 0.0117 & \underline{0.6496} & \underline{0.0260} & \textbf{0.0122} & \underline{0.3201}  & \underline{0.0605} & \underline{0.0212} & \textbf{0.5595}\\
    \bottomrule
    \end{tabular}
    \vspace{-1.5em}
    \end{table*}
    As for specific supervision, we apply a relative supervision for each predicted pose pair. Formally, let $S_{i\rightarrow j}$ denote the relative transform at time $t$. $R_{i\rightarrow j}$ and $t_{i\rightarrow j}$ respectively represents the rotation matrix and transform vector. The camera loss is defined as:
    \begin{equation}
    \mathcal{L}_{\text{cam}} = \frac{1}{T(T-1)}\sum_{i\neq j}(\theta_{\hat{R}_{i\rightarrow j},R_{i\rightarrow j}}+\|\hat{t}_{i\rightarrow j}-t_{i\rightarrow j}\|),
    \end{equation}
    where $T$ is the total number of frames in the sequence. 
    
    Additionally, this loss is applied differently to the original and duplicated camera tokens. For the original camera tokens, when computing losses over relative transforms from earlier to later frames, gradients are detached for tokens from the earlier timestamps to prevent back-propagation through the entire temporal chain. In contrast, for the duplicated camera tokens, we retain full gradient flow across all temporal relations.

    \begin{figure*}[!h]
      \centering
      \vspace{-1em}
      \includegraphics[width=\linewidth]{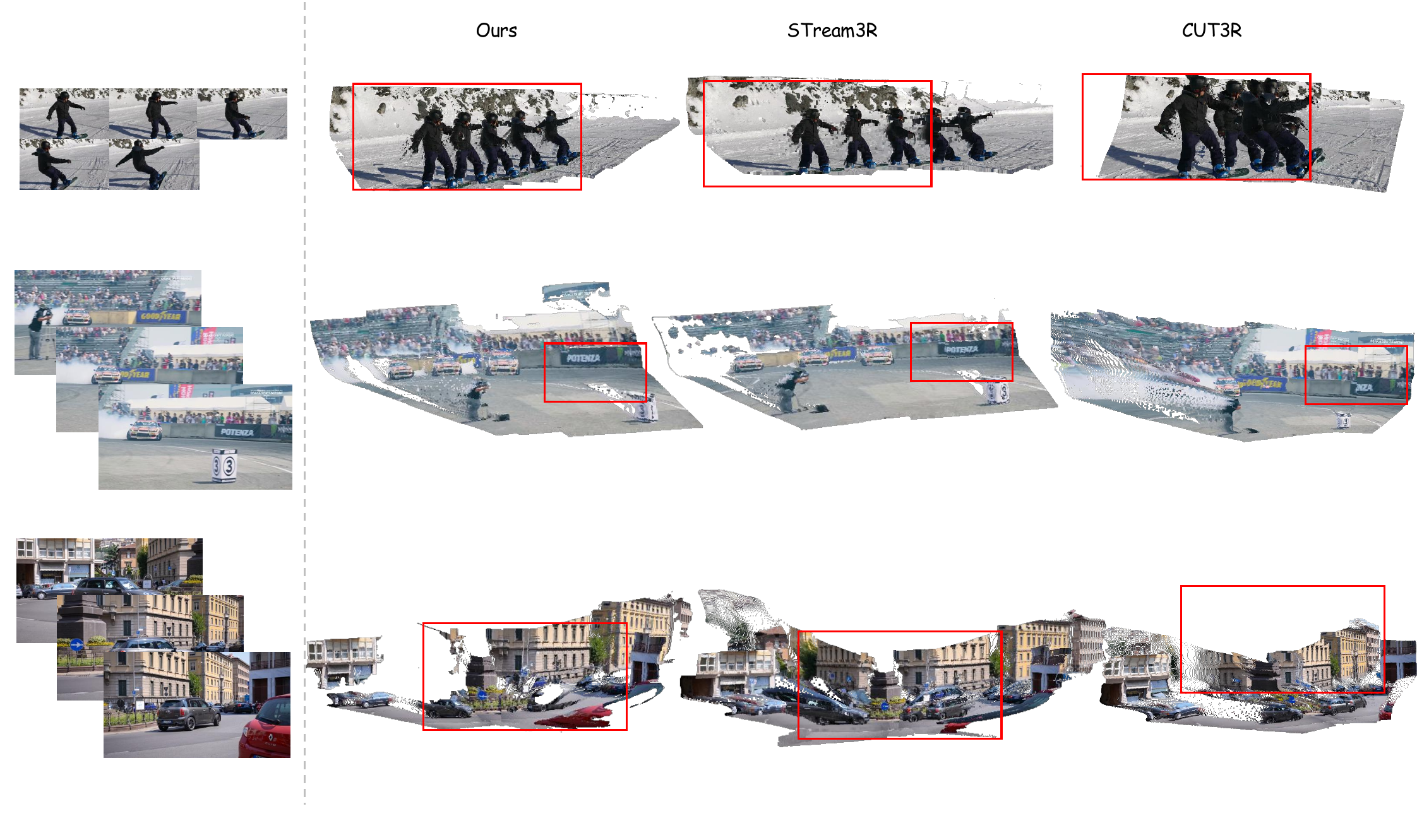}
      \caption{\textbf{Qualitative Comparison of our Stream Model with Other Methods.} MoRe strikes an optimal balance between reconstruction quality and computational efficiency, delivering high-fidelity results at competitive inference speeds.}
      \label{fig:visualization_stream}
      \vspace{-1.5em}
    \end{figure*}
\section{Experiments}
\paragraph{Datasets.} Our model is trained on a large and diverse collection of datasets encompassing both static and dynamic scenes. Specifically, we utilize Dynamic Replica~~\cite{karaev2023dynamicstereo}, PointOdyssey~~\cite{zheng2023pointodyssey}, Spring~~\cite{mehl2023spring}, Virtual KITTI~~\cite{cabon2020virtual}, TartanAir~~\cite{wang2020tartanair}, Co3Dv2~~\cite{reizenstein2021common}, ScanNet~~\cite{dai2017scannet}, BlendedMVS~~\cite{yao2020blendedmvs}, Hypersim~~\cite{roberts2021hypersim}, ARKitScenes~~\cite{baruch2021arkitscenes}, Waymo~~\cite{sun2020scalability}, and OmniWorld-Game~~\cite{zhou2025omniworld}. These datasets jointly cover a wide variety of indoor and outdoor environments, object categories, lighting conditions, and motion patterns. To ensure a balanced training distribution, datasets with fewer sequences are duplicated proportionally.
    \subsection{Camera Pose Estimation}
    \begin{table*}
    \vspace{-1em}
    \centering
    \footnotesize
    \setlength{\tabcolsep}{6pt}
    \caption{Video Depth Estimation on Sintel~\cite{butler2012naturalistic}, TUM-dynamics~\cite{sturm2012benchmark}, Bonn~\cite{palazzolo2019refusion} and kitti~\cite{geiger2013vision}.}
    \label{tab:video_depth_estimation}
    \begin{tabular}{c c cc|cc|cc|cc}
    \toprule
     & &\multicolumn{2}{c}{\textbf{Sintel}} & \multicolumn{2}{c}{\textbf{Bonn}} & \multicolumn{2}{c}{\textbf{TUM-dynamics}} & \multicolumn{2}{c}{\textbf{kitti}} \\
    \cmidrule(lr){3-4} \cmidrule(lr){5-6} \cmidrule(lr){7-8} \cmidrule(lr){9-10}
     \textbf{Method} & \textbf{type} & Abs Rel$\downarrow$  & $\delta<1.25\uparrow$ & Abs Rel$\downarrow$  & $\delta<1.25\uparrow$ & Abs Rel$\downarrow$ & $\delta<1.25\uparrow$ & Abs Rel$\downarrow$ & $\delta<1.25\uparrow$ \\
    \midrule
    Flare~\cite{zhang2025flare} & FA & 0.729 & 0.336 & 0.116 & 0.897 & 0.152 & 0.790 & 0.356 & 0.570 \\
    MapAnything~\cite{keetha2025mapanything} & FA & 0.632 & 0.461 & 0.125 & 0.859 & 0.241 & 0.813 & 0.277 & 0.742\\
    VGGT~\cite{wang2025vggt} & FA & \underline{0.387} & \underline{0.584} & \textbf{0.055} & \underline{0.970} & \underline{0.132} & \underline{0.901}  & \underline{0.073} & \underline{0.961}\\
    MoRe(FA) & FA &  \textbf{0.335} & \textbf{0.645} & \textbf{0.055} & \textbf{0.971} & \textbf{0.120} & \textbf{0.902}  & \textbf{0.066} & \textbf{0.967}\\
    \midrule
    Spann3R~\cite{wang20243d} & Streaming & 0.740 & 0.364 & 0.229  & 0.729 & 0.264 & 0.645 & 0.422 & 0.358   \\
    CUT3R~\cite{wang2025continuous} & Streaming & 1.363 & 0.434 & 0.076 & 0.949 & 0.160 & 0.856 & 0.124 & 0.873   \\
    StreamVGGT~\cite{zhuo2025streaming} & Streaming & 0.698 & 0.591 & 0.058 & \textbf{0.972} & \underline{0.156} & \underline{0.889} & 0.173 & 0.722   \\
    Wint3R~\cite{li2025wint3r} & Streaming & 0.830 & 0.537 & 0.071 & 0.912 & 0.177 & 0.808 & 0.201 & 0.601   \\
    Stream3R~\cite{lan2025stream3r} & Streaming & \underline{0.397} & \underline{0.632} & \underline{0.070} & 0.952 & \textbf{0.151} & \textbf{0.898} & \underline{0.079} & \underline{0.949}   \\
    MoRe & Streaming &  \textbf{0.254} & \textbf{0.637} & \textbf{0.068} & \underline{0.961} & 0.173 & 0.868 & \textbf{0.072} & \textbf{0.966} \\
    \bottomrule
    \end{tabular}
    \vspace{-1em}
    \end{table*}
    Following previous works~~\cite{wang2025continuous, lan2025stream3r}, we evaluate our predicted camera poses on Sintel~\cite{butler2012naturalistic}, TUM-dynamics~\cite{sturm2012benchmark}, Bonn~\cite{palazzolo2019refusion}, and ScanNet~\cite{dai2017scannet}. The first three are dynamic datasets containing a large proportion of moving objects. Among them, TUM-dynamics and Bonn are real-world datasets with random camera jitter and complex motion patterns, posing significant challenges to static reconstruction methods. These properties make them ideal for evaluating camera pose estimation in 4D reconstruction scenarios. Notably, none of these dynamic datasets are seen during training, demonstrating the zero-shot generalization ability of our model.In contrast, ScanNet is a static indoor dataset, which helps verify that our model not only performs well in dynamic 4D settings but also maintains strong capability on static reconstruction tasks. Our evaluation metrics include Absolute Translation Error (ATE), Relative Translation Error ($\text{RPE}_\text{trans}$), and Relative Rotation Error ($\text{RPE}_\text{rot}$), computed post Sim(3) alignment with ground truth poses. Quantitative results are summarized in \cref{tab:pose_estimation}. Among full-attention methods, our approach achieves comparable performance to the state-of-the-art $\pi^3$~\cite{wang2025pi}, despite being trained with significantly less data. Moreover, our method consistently outperforms streaming-based approaches, highlighting the effectiveness of our attention-forcing strategy in disentangling dynamic motion from static structure. This demonstrates that explicitly guiding attention during training leads to more robust and accurate camera pose estimation under challenging dynamic scenes.

    \subsection{Video Depth Estimation}
    We evaluate depth prediction on four widely used benchmarks, including Sintel, Bonn, TUM, and kitti, which collectively cover diverse scenarios, including both synthetic and real-world data, as well as indoor and outdoor environments. This provides a comprehensive evaluation of the model’s generalization ability across different domains.To quantify performance, we adopt standard metrics: the Absolute Relative Error (Abs-Rel), which measures the average proportional deviation between predicted and ground-truth depths, and $\delta_{1.25}$ accuracy, representing the percentage of predictions within a multiplicative factor of 1.25 from the ground truth. Following prior works~~\cite{wang2025continuous, lan2025stream3r}, all predicted depth maps are aligned to the ground truth via a scale-only transformation before computing metrics.As shown in \cref{tab:video_depth_estimation}, our model achieves consistently strong results across all benchmarks. It performs competitively among full-attention methods and clearly surpasses existing streaming-based approaches, demonstrating its strong capability in monocular video depth estimation. Combined with the accurate camera pose estimation discussed earlier, these results validate that MoRe is well-suited for unified and robust 4D reconstruction.

    \subsection{Ablation Study}
    We perform ablation studies to investigate how different components of our model contribute to its overall performance. In particular, we examine the impact of the attention forcing mechanism, grouped causal attention and BA-like refinement under consistent training configurations.
    \vspace{-1em}
    \paragraph{Attention Forcing}
    \begin{table}[t]
    \vspace{-1em}
    \centering
    \footnotesize
    \setlength{\tabcolsep}{0.5pt}
    \caption{Ablation on Camera Pose Estimation}
    \label{tab:attention_forcing_ablation}
    \begin{tabular}{c ccc ccc}
    \toprule
     & \multicolumn{3}{c}{\textbf{Sintel}} & \multicolumn{3}{c}{\textbf{TUM-dynamics}}   \\
     \cmidrule(lr){2-4} \cmidrule(lr){5-7}
     \textbf{Method} & ATE$\downarrow$ & RPE$_\text{trans}$$\downarrow$ & RPE$_\text{rot}$$\downarrow$ & ATE$\downarrow$ & RPE$_\text{trans}$$\downarrow$ & RPE$_\text{rot}$$\downarrow$ \\
    \midrule
    w/o attention forcing & 0.163 & 0.092 & 0.660 & 0.028 & 0.014 & 0.329  \\
    w/o BA-like refinement & 0.155 & 0.085 & 0.619 & 0.027 & 0.014 & 0.321   \\
    \midrule
    full  & \textbf{0.147} & \textbf{0.082} & \textbf{0.616} & \textbf{0.026} & \textbf{0.013} & \textbf{0.320}\\
    \bottomrule
    \end{tabular}
    \vspace{-2em}
    \end{table}
    To validate of the proposed attention forcing strategy, we train two variants of our model: one with attention forcing and one without. Both models share the identical architecture, training schedule, and data configuration. We report their camera pose estimation results on the Sintel~\cite{butler2012naturalistic} and TUM-Dynamics~\cite{sturm2012benchmark} in \cref{tab:attention_forcing_ablation}. As shown in the table, introducing the attention forcing mechanism significantly improves the accuracy of camera pose estimation. This demonstrates that explicitly guiding the attention map toward static regions helps the model better disentangle dynamic motion from static structures. 
    \vspace{-1em}
    \paragraph{Grouped Causal Attention}
    \begin{table}[h]
    \vspace{-1em}
    \centering
    \footnotesize
    \setlength{\tabcolsep}{0.6pt}
    \caption{Ablation on Video Depth Estimation}
    \label{tab:causal_attention_ablation}
    \begin{tabular}{c cc cc cc}
    \toprule
     & \multicolumn{2}{c}{\textbf{Sintel}} & \multicolumn{2}{c}{\textbf{Bonn}} & \multicolumn{2}{c}{\textbf{kitti}}  \\
     \cmidrule(lr){2-3} \cmidrule(lr){4-5} \cmidrule(lr){6-7}
     \textbf{Method} & Abs Rel$\downarrow$ & $\delta<1.25\uparrow$ & Abs Rel$\downarrow$ & $\delta<1.25\uparrow$ & Abs Rel$\downarrow$ & $\delta<1.25\uparrow$\\
    \midrule
    w/o GCA & 0.277	& 0.592 & 0.070 & 0.953  & 0.079 & 0.949\\
    \midrule
    w/  & \textbf{0.254} & \textbf{0.637} & \textbf{0.068} & \textbf{0.961} & \textbf{0.072} & \textbf{0.966}\\
    \bottomrule
    \end{tabular}
    \vspace{-2em}
    \end{table}
    To evaluate the effectiveness of our grouped causal attention, we compare it against a variant that adopts the standard causal attention for both training and inference. The baseline model is implemented with FlashAttention, and all other configurations are kept identical to ensure a fair comparison. We report depth estimation metrics on Sintel~\cite{butler2012naturalistic}, Bonn~\cite{palazzolo2019refusion} and kitti~\cite{geiger2013vision}, as summarized in \cref{tab:causal_attention_ablation}, where GCA refers to Grouped Causal Attention. Results show that our grouped causal attention consistently improves performance across datasets. This design enables each image token to attend to others within the same frame while maintaining causal temporal dependency, effectively enhancing both temporal reasoning and spatial consistency. 
    \vspace{-1em}
    \paragraph{BA-like refinement}
    We ablate the BA-like refinement by removing the duplicated camera tokens used for post-process refinement. Results on Sintel~\cite{butler2012naturalistic} and TUM-dynamics~\cite{sturm2012benchmark} are shown in \cref{tab:attention_forcing_ablation}. The variant without refinement shows noticeably higher translation and rotation errors, confirming that our BA-like refinement effectively improves pose accuracy and temporal consistency in streaming reconstruction.
    
\section{Conclusion}
We introduced MoRe, a feed-forward network for dynamic 4D reconstruction from monocular videos that effectively tackles challenges arising from moving objects and camera motion. By incorporating an attention-forcing strategy, MoRe explicitly disentangles dynamic motion from static scene geometry without requiring explicit motion priors during inference. Our streaming inference mechanism which combines grouped causal attention with a lightweight BA–like refinement, achieves efficient and temporally coherent reconstructions. 
{
    \small
    \bibliographystyle{ieeenat_fullname}
    \bibliography{main}
}

\clearpage
\setcounter{page}{1}
\maketitlesupplementary
\section{Training Details}
We train our model using the AdamW optimizer~~\cite{loshchilov2017decoupled} to minimize the overall loss function. The training is conducted for 100K iterations with a learning rate scheduler that includes a warm-up phase and a peak learning rate of $1\times10^{-6}$. At each iteration, we randomly sample 2–24 frames from each sequence with a temporal interval of 1–5. The input images are resized such that the longer side is fixed to 518 pixels, while the shorter side is randomly scaled by a factor of 0.8–1.2 for data augmentation.
Training is performed on 64 NVIDIA A800 GPUs for approximately two days. We adopt bfloat16 precision and gradient checkpointing to reduce memory consumption and enable efficient large-scale training.
\section{Motion Mask Extraction}
    \subsection{Data Preparation}
    
    Most existing datasets lack reliable motion-mask annotations, making it difficult to obtain high-quality supervision for dynamic scene understanding. To address this issue, we propose a robust motion-mask extraction pipeline. Given raw images, we first apply SAM2~\cite{ravi2024sam} to obtain semantic segmentation masks. The ego flow is computed from ground-truth camera poses and intrinsics, while SEA-RAFT~\cite{wang2024sea} predicts the optical flow.
    
    For each semantic region $S_k$, we compute the average flow discrepancy:
    \begin{equation}
    d_k = \frac{1}{|S_k|}
    \sum_{(i,j)\in S_k}
    \left\| \mathbf{F}^{\text{pred}}(i,j) - \mathbf{F}^{\text{ego}}(i,j) \right\|_2 .
    \end{equation}
    
    A semantic region is considered moving if its discrepancy exceeds a statistical threshold:
    \begin{equation}
    d_k > \mu_d + 2\sigma_d .
    \end{equation}
    
    Finally, the motion mask $M(u,v)$ is defined as:
    \begin{equation}
    M(u,v) =
    \begin{cases}
    1, & \text{if } (u,v)\in S_k \ \text{and}\ d_k > \mu_d + 2\sigma_d, \\[2mm]
    0, & \text{otherwise.}
    \end{cases}
    \end{equation}
    \subsection{Qualitative Results}
    \begin{figure}[!h]
      \centering
      \includegraphics[width=\linewidth]{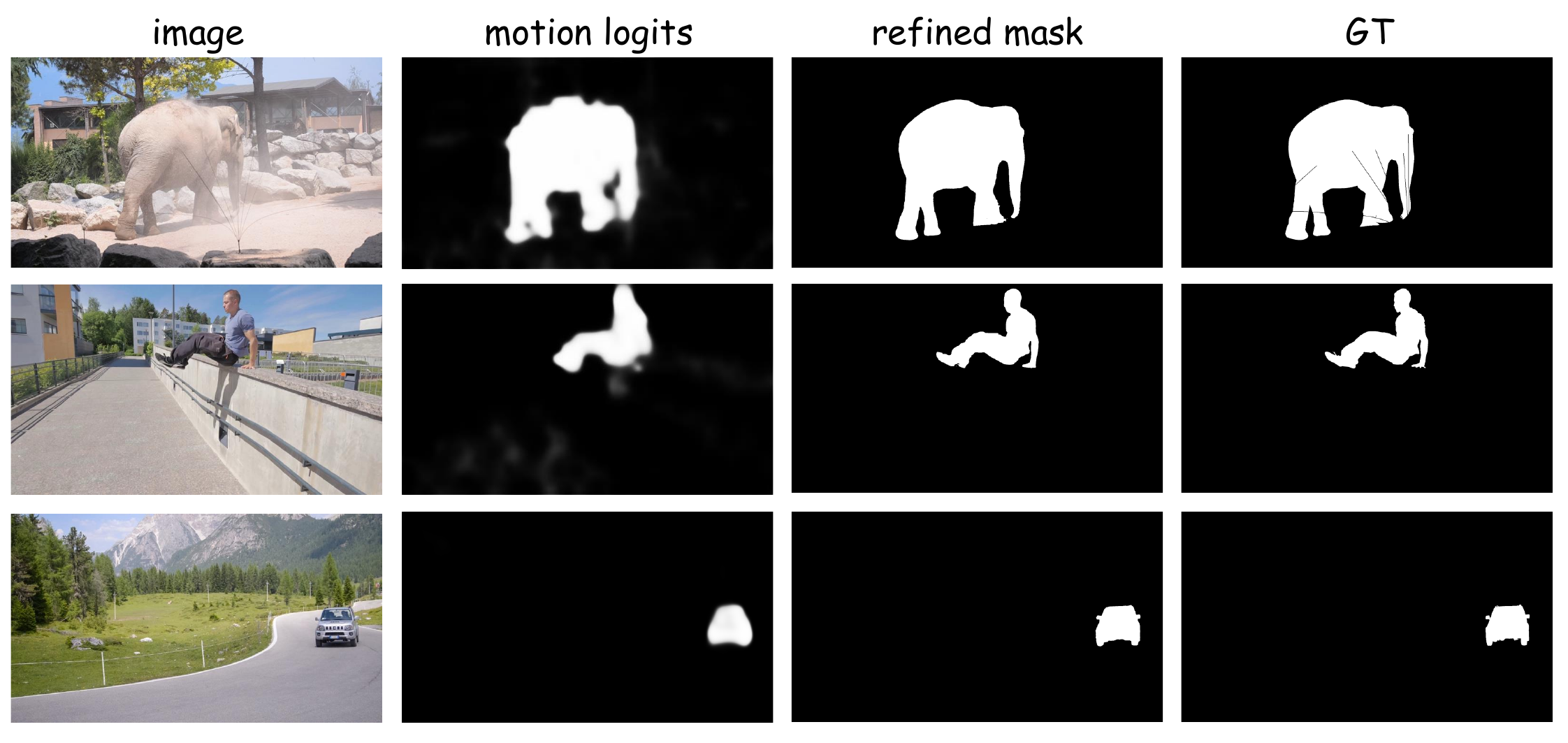} %
      \caption{\textbf{Qualitative Results of Motion Mask Extraction.} Our method robustly captures moving objects across diverse scenes and objects.}
      \label{fig:motion masks visualization}
    \end{figure}
    We evaluate our method on the DAVIS~\cite{pont20172017} dataset. For visualization, we present both the raw outputs of our model and a refined version obtained by applying the image-level predictor of SAM2~\cite{ravi2024sam}. As shown in \cref{fig:motion masks visualization}, our approach consistently produces accurate motion-object segmentation across diverse scenes and object categories. After simple post-processing, the generated motion masks are sufficiently clean and robust to be directly used in downstream tasks such as dynamic scene reconstruction, moving-object removal, and motion-aware 4D generation.
\section{Stream Inference}
    \subsection{Implementation Details}
    Streaming generation has been widely adopted in large language models and related multi-modal systems to reduce latency and computational cost~\cite{touvron2023llama,xiao2023efficient,ning2024inf}. Inspired by this paradigm, we introduce streaming and causal attention mechanisms into MoRe, enabling real-time, constant-latency generation with image-wise KV caching. This design effectively avoids redundant computation by reusing the stored key–value pairs from previous steps. In addition, we incorporate a window-sliding strategy to prevent unbounded growth of the KV cache.
    
    We employ two streamers: an input streamer for continuous image ingestion and an output streamer for delivering predictions. In the work flow, each new image enters an infinite decoding loop, where its hidden states are concatenated with cached keys and values -- optionally applying window sliding -- before passing through the stack of N transformer layers. The updated prediction is then immediately emitted through the output streamer, enabling continuous and low-latency streaming outputs.
    \subsection{Efficiency Test}
    \begin{table}[t]
    \setlength{\tabcolsep}{10pt}
    \centering
    \caption{Inference speed comparison (FPS), tested on KITTI~\cite{geiger2013vision}.}
    \label{tab:fps_comparison}
    \begin{tabular}{l c}
    \toprule
    \textbf{Method} & \textbf{FPS} \\
    \midrule
    VGGT~\cite{wang2025vggt}                          & 7.32  \\
    Spann3R~\cite{wang20243d} \shortstack{(224×224)}  & 13.55 \\
    CUT3R~\cite{wang2025continuous} & 16.58 \\
    Stream3R$^\alpha$~\cite{lan2025stream3r}             & 23.48 \\
    Stream3R$^\beta$~\cite{lan2025stream3r}              & 12.95 \\
    Stream3R$^\beta$-W[5]~\cite{lan2025stream3r} \shortstack{(window=5)} & 32.93 \\
    \midrule
    MoRe                 & 30.09 \\
    \bottomrule
    \end{tabular}
    \end{table}
    \begin{figure*}[!t]
      \centering
      \includegraphics[width=\linewidth]{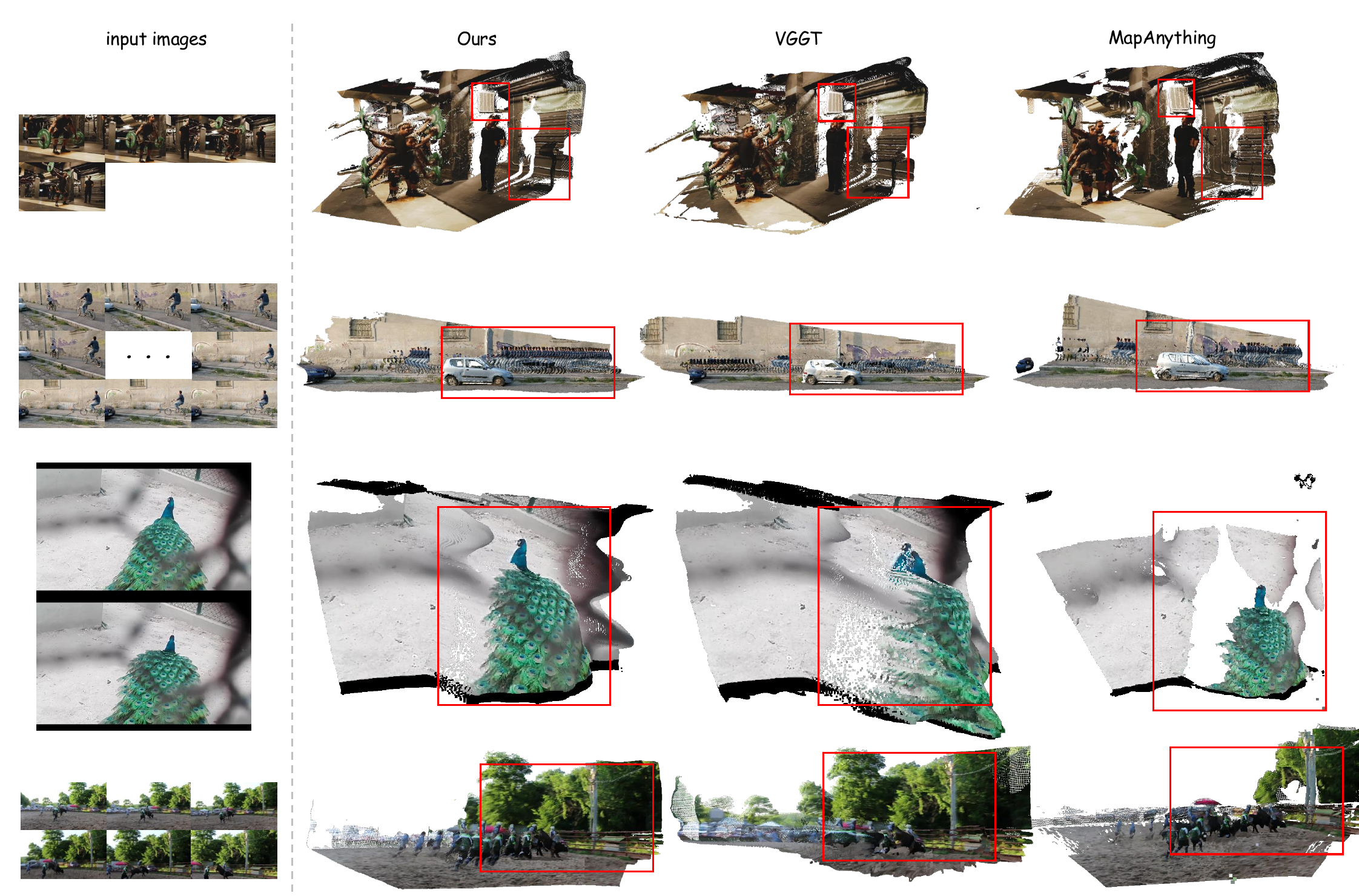}
      \caption{\textbf{Qualitative Comparison of Our Model with Other Methods.} We present an extensive set of visual results showcasing reconstruction quality, motion handling, and robustness under challenging dynamic scenarios.}
      \label{fig:visualization_full}
    \end{figure*}
    We evaluate the inference speed of our method on the KITTI dataset at a resolution of 512×144 using an NVIDIA A800 GPU, ensuring consistency across all compared approaches except for Spann3R~\cite{wang20243d}, which processes Stream inputs at a resolution of 224×224. The performance of other baselines follows the Stream3R\cite{lan2025stream3r} evaluation results. In addition, we report the FPS of our model employing a sliding window attention mechanism with a window size of 5. As shown in \cref{tab:fps_comparison}, our method achieves inference speeds within the fastest tier among all evaluated methods, outperforming most baselines while maintaining competitive reconstruction accuracy. This demonstrates that our approach provides an excellent trade-off between speed and performance, making it highly suitable for real-time 4D reconstruction systems and applications.
    \subsection{Qualitative Results}
    To qualitatively evaluate the reconstruction quality of our approach, we further visualize the dynamic 4D scenes reconstructed from monocular video sequences. As shown in \cref{fig:visualization_full}, our method effectively captures both static scene geometry and dynamic object motion with high fidelity and temporal coherence. The detailed geometry and consistent motion trajectories demonstrate the robustness of our model in handling complex dynamic environments. These visual results further validate the effectiveness of our approach for practical 4D reconstruction applications. In addition to the stream inference, we also provide more examples of the full attention model. For some methods, slight deviations in the rendered viewpoint occur because their reconstructed point clouds have different scales.
\section{Motion Aligned Attention}
\subsection{Quantitative Results}
We further evaluate our model on the Co3Dv2~\cite{reizenstein2021common} dataset to verify its capability in static scene reconstruction, and we additionally compare against a broader range of baselines. The results are summarized in \cref{tab:comparison}. As discussed in the main text, our full-attention variant achieves the best overall performance and surpasses all baselines, including the state-of-the-art $\pi^3$ method. These results demonstrate that, although our architecture and training strategy are primarily designed for modeling dynamic scenes and suppressing motion-induced ambiguities, the model retains excellent reconstruction accuracy on purely static scenarios. This highlights the strong robustness and generalization ability of our approach: the motion-aware design does not compromise performance when no motion is present, and instead enables the model to effectively capture both dynamic and static structural cues in a unified framework.

\begin{table}[t]
\centering
\caption{Camera Pose Estimation Comparison on Co3Dv2~\cite{reizenstein2021common}.}
\label{tab:comparison}
\small
\begin{tabular}{l|ccc}
\toprule
{\textbf{Method}} &
\multicolumn{3}{c}{\textbf{Co3Dv2}} \\
 \cmidrule(lr){2-4}
& RRA@30$\uparrow$ & RTA@30$\uparrow$ & AUC@30$\uparrow$ \\
\midrule
Fast3R~\cite{yang2025fast3r}     & 97.49 & 91.11 & 73.43 \\
CUT3R~\cite{wang2025continuous}       & 96.19 & 92.69 & 75.82 \\
FLARE~\cite{zhang2025flare}       & 96.38 & 93.76 & 73.99 \\
VGGT~\cite{wang2025vggt}         & 98.96 & 97.13 & 88.59 \\
$\pi^3$~\cite{wang2025pi} & 99.05 & 97.33 & 88.41 \\
\midrule
MoRe & \textbf{99.49} & \textbf{98.11} & \textbf{91.42}\\
\bottomrule
\end{tabular}
\end{table}
\subsection{Visualization}
\begin{figure}[!h]
      \centering
      \includegraphics[width=\linewidth]{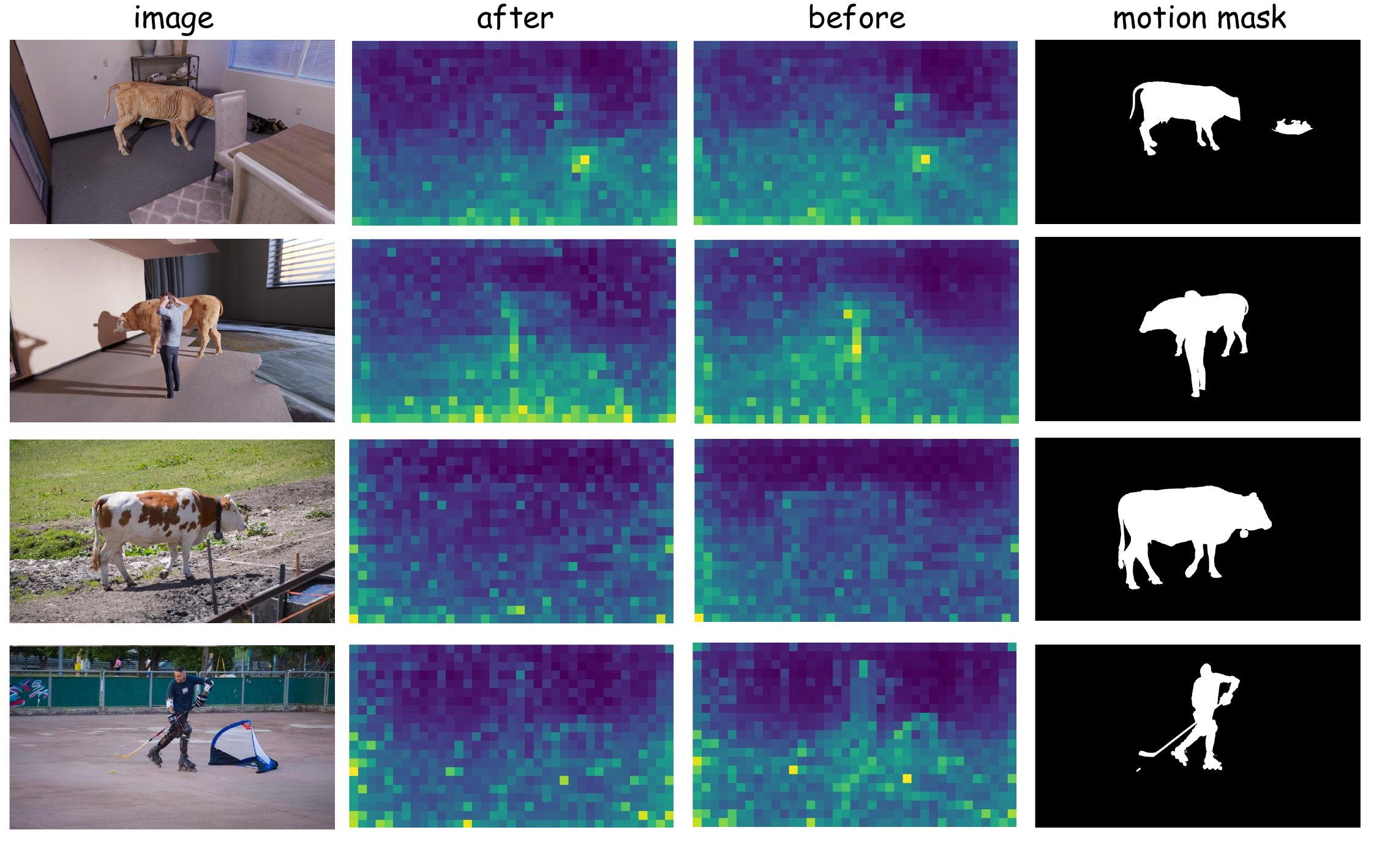} %
      \caption{\textbf{Attention Map Comparison.} We visualize the attention map on Dynamic Replica~\cite{sucar2025dynamic} and DAVIS~\cite{pont20172017} dataset. Our motion-aligned training suppresses undesired attention from camera tokens to dynamic objects, yielding cleaner and more structured attention patterns.}
      \label{fig:attn map comparison}
    \end{figure}
To better illustrate the effectiveness of our motion-aligned training strategy, we visualize and compare the attention weight maps of camera token before and after training. Specifically, we select the last attention layer of our model and compute the average attention weight across all heads to obtain a stable and interpretable heatmap representation. As shown in \cref{fig:attn map comparison}, the attention distribution becomes significantly more structured. The attention weight from camera tokens to dynamic objects is notably suppressed, while attention toward static regions becomes more concentrated and semantically coherent. This indicates that the model has learned to reserve camera tokens for representing stable scene information, preventing dynamic content from leaking into the global representation. The resulting separation leads to cleaner latent features, more stable motion reasoning, and ultimately more accurate 4D reconstruction. These observations validate that our training strategy effectively regularizes the attention behavior and enforces the intended representational roles of different token types.
\subsection{Loss Design}
We initially explored divergence-based formulations, such as applying a KL divergence to align the attention distribution to the motion-score distribution. While this approach appears principled, it implicitly normalizes attention into a probability distribution, which tends to introduce an undesirable inductive bias in static scenes.
In static regions, the correct behavior should allow attention weights to remain largely unconstrained, whereas KL-based losses force all tokens to contribute to a normalized distribution even when no motion exists, leading to degraded performance. The constant $C$ serves as a neutral baseline representing the default attention level.
During training, tokens with high motion scores ($\hat{a}_i$ large) are encouraged to deviate from this baseline, while tokens associated with static content ($\hat{a}_i \approx 0$) receive minimal gradient updates. This yields a motion-adaptive behavior that avoids imposing constraints where no motion is present. The multiplicative term ($\alpha_i - C)\,\hat{a}_i$ acts as a gating mechanism, in which motion-relevant tokens receive stronger supervision and motion-irrelevant tokens are softly ignored. This formulation provides flexibility and avoids the normalization issues inherent to divergence losses. We conducted ablations comparing the proposed loss with a KL-based alternative. As shown in \cref{tab:loss_function_ablation}, the KL formulation performs worse in both static and low-motion scenarios, confirming that the proposed motion-gated design better matches the nature of the task and provides more stable training behavior.
\begin{table}[t]
    \vspace{-1em}
    \centering
    \footnotesize
    \setlength{\tabcolsep}{2pt}
    \caption{Ablation on Loss Function for Motion Alignment.}
    \label{tab:loss_function_ablation}
    \begin{tabular}{c ccc ccc}
    \toprule
     & \multicolumn{3}{c}{\textbf{Sintel}} & \multicolumn{3}{c}{\textbf{TUM-dynamics}}   \\
     \cmidrule(lr){2-4} \cmidrule(lr){5-7}
     \textbf{Method} & ATE$\downarrow$ & RPE$_\text{trans}$$\downarrow$ & RPE$_\text{rot}$$\downarrow$ & ATE$\downarrow$ & RPE$_\text{trans}$$\downarrow$ & RPE$_\text{rot}$$\downarrow$ \\
    \midrule
    w/ KL loss & 0.185 & 0.084 & 0.707  & 0.029 & 0.015 & 0.350   \\
    \midrule
    Ours  & \textbf{0.147} & \textbf{0.082} & \textbf{0.616} & \textbf{0.026} & \textbf{0.013} & \textbf{0.320}\\
    \bottomrule
    \end{tabular}
    \vspace{-1em}
    \end{table}
\section{Limitations}
Despite demonstrating strong performance in dynamic scenes, our method has several limitations. First, although our method achieves strong results in dynamic scene modeling, it heavily depends on the accuracy and quality of motion mask annotations. Since the motion masks provide critical supervision to distinguish moving regions from static background, any errors, noise, or inconsistencies in these masks can propagate through the training process, leading to degraded reconstruction quality and less reliable motion reasoning. This reliance poses a limitation, especially when high-quality motion mask labels are unavailable or difficult to obtain in real-world scenarios. Future work could explore more robust or self-supervised techniques to mitigate the impact of imperfect motion supervision and reduce dependency on manual or heuristic mask extraction. Second, while the feed-forward architecture enables efficient and real-time inference, it may struggle to capture very long-term temporal dependencies and complex dynamic interactions that extend beyond the modeled temporal window. Third, the model may exhibit reduced robustness in scenes with extremely fast or non-rigid motions, where motion patterns are highly irregular and difficult to disentangle. In addition, our model can fail in heavily motion-blurred scenarios, where rapid camera movement or fast object motion leads to severely degraded visual cues. In such cases, attention alignment becomes unreliable, causing inaccurate depth, unstable poses, or distorted geometry. Lastly, our current approach does not explicitly handle occlusions or severe appearance changes over time, which can lead to artifacts or inconsistencies in the reconstructed 4D scenes. Addressing these challenges is an important direction for future research.


\end{document}